\documentclass[11pt,letterpaper]{article}
\usepackage{times}
\usepackage{latexsym}

\usepackage{hyperref}
\usepackage{basecommon} 

\usepackage{naaclhlt2016}

\usepackage{balance}

\naaclfinalcopy 
\def\naaclpaperid{***} 

\title{Sentence-Level Grammatical Error Identification \\as Sequence-to-Sequence Correction}

\author{Allen Schmaltz \And Yoon Kim \And Alexander M. Rush \And Stuart M. Shieber \AND
Harvard University \\
{\tt \footnotesize \{schmaltz@fas,yoonkim@seas,srush@seas,shieber@seas\}.harvard.edu}
}

\DeclareMathOperator{\softmax}{softmax}
\DeclareMathOperator{\relu}{ReLU}
\DeclareMathOperator{\lstm}{LSTM} 

\newcommand{\xvec}{\mathbf{x}}

\newcommand{\cvec}{\mathbf{c}}
\newcommand{\zvec}{\mathbf{z}}
\newcommand{\svec}{\mathbf{s}}
\newcommand{\tvec}{\mathbf{t}}

\newcommand{\Uvec}{\mathbf{U}}

\newcommand{\Wvec}{\mathbf{W}}
\newcommand{\hvec}{\mathbf{h}}

\newcommand{\bvec}{\mathbf{b}}

\date{}

\begin{document}

\maketitle

\begin{abstract}
 We demonstrate that an attention-based encoder-decoder model can be used for sentence-level grammatical error identification for the Automated Evaluation of Scientific Writing (AESW) Shared Task 2016. The attention-based encoder-decoder models can be used for the generation of corrections, in addition to error identification, which is of interest for certain end-user applications. We show that a character-based encoder-decoder model is particularly effective, outperforming other results on the AESW Shared Task on its own, and showing gains over a word-based counterpart. Our final model---a combination of three character-based encoder-decoder models, one word-based encoder-decoder model, and a sentence-level CNN---is the highest performing system on the AESW 2016 binary prediction Shared Task.
\end{abstract}

\section{Introduction}

The recent confluence of data availability and strong sequence-to-sequence learning algorithms has the potential to lead to practical tools for writing support. Grammatical error identification is one such application of potential utility as a component of a writing support tool. Much of the recent work in grammatical error identification and correction has made use of hand-tuned rules and features that augment data-driven approaches, or individual classifiers for human-designated subsets of errors. Given a large, annotated dataset of scientific journal articles, we propose a fully data-driven approach for this problem, inspired by recent work in neural machine translation and more generally, sequence-to-sequence learning \cite{SutskeverEtAl-2014-SequenceToSequence,BahdanauEtAl-2014-NMTAttention,ChoEtAl-2014-RNNEncDecNMT}.

The Automated Evaluation of Scientific Writing (AESW) 2016 dataset is a collection of nearly 10,000 scientific journal articles (over 1 million sentences) published between 2006 and 2013 and annotated with corrections by professional, native English-speaking editors. The goal of the associated  AESW Shared Task is to identify whether or not a given unedited source sentence was corrected by the editor (that is, whether a given source sentence has one or more grammatical errors, broadly construed).

This system report describes our approach and submission to the AESW 2016 Shared Task, which establishes the current highest-performing public baseline for the binary prediction task. Our primary contribution is to demonstrate the utility of an attention-based encoder-decoder model for the binary prediction task. We also provide evidence of tangible performance gains using a character-aware version of the model, building on the character-aware language modeling work of \newcite{KimEtAl-2016-CharLM}.
In addition to sentence-level classification, the models are capable of intra-sentence error identification and the generation of possible corrections. We also obtain additional gains by using an
ensemble of a generative encoder-decoder and a discriminative CNN classifier.

\section{Background}

Recent work in natural language processing has shown strong results in sequence-to-sequence transformations using recurrent neural network models \cite{ChoEtAl-2014-RNNEncDecNMT,SutskeverEtAl-2014-SequenceToSequence}. Grammar correction and error identification can be cast as a sequence-to-sequence translation problem, in which an unedited (\textit{source}) sentence is ``translated'' into a corrected (\textit{target}) sentence in the same language. Using this framework, sentence-level error identification then simply reduces to an equality check between the source and target sentences.

The goal of the AESW shared task is to identify whether a particular sentence needs to be edited (contains a ``grammatical'' error, broadly construed\footnote{Some insertions and deletions in the shared task data represent stylistic choices, not all of which are necessarily recoverable given the sentence or paragraph context. For the purposes here, we refer to all such edits as ``grammatical'' errors.}). The dataset consists of sentences taken from academic articles annotated with corrections by professional editors. Annotations are described via insertions and deletions, which are marked with start and end tags. Tokens to be deleted are surrounded with the deletion start tag~$\mathsf{ \textless del\textgreater}$ and the deletion end tag~$\mathsf{ \textless /del\textgreater}$ and tokens to be inserted are surrounded with the insertion start tag~$\mathsf{ \textless ins\textgreater}$ and the insertion end tag~$\mathsf{ \textless /ins\textgreater}$. Replacements (as shown in Figure~\ref{fig:architecture}) are represented as deletion-insertion pairs. Unlike the related CoNLL-2014 Shared Task \cite{NgEtAl-2014-SharedTask2014} data, errors are not labeled with fine-grained types (article or determiner error, verb tense error, etc.).

More formally, we assume a vocabulary $\mcV$ of natural language word types (some of which have orthographic errors) and a set $\mcQ=\{\mathsf{ \textless ins\textgreater},\mathsf{ \textless /ins\textgreater},\mathsf{ \textless del\textgreater},\mathsf{ \textless /del\textgreater} \}$ of annotation tags. 
Given a sentence $\mathbf{s} = [s_1, \dots, s_I]$, where $s_i \in \mcV$ is the $i$-th token of the sentence of length $I$, we seek to predict whether or not the gold, annotated target sentence $\mathbf{t} = [t_1, \dots, t_J]$, where $t_j \in \mcQ \cup \mcV$ is the $j$-th token of the annotated sentence of length $J$, is identical to $\mathbf{s}$.  
We are given both $\mathbf{s}$ and $\mathbf{t}$ for supervised training. At test time, we are only given access to sequence $\mathbf{s}$. We learn to predict sequence $\mathbf{t}$.

Evaluation of this binary prediction task is via the $F_1$-score, where the positive class is that indicating an error is present in the sentence (that is, where $\mathbf{s} \neq \mathbf{t}$)\footnote{The 2016 Shared Task also included a probabilistic estimation track. We leave for future work the adaptation of our approach to that task.}.

Evaluation is at the sentence level, but the paragraph-level context for each sentence is also provided. The paragraphs, themselves, are shuffled so that full article context is not available. A coarse academic field category is also provided for each paragraph. Our models described below do not make use of the paragraph context nor the field category, and they treat each sentence independently.

Further information about the task is available in the Shared Task report \cite{daudaravicius-banchs-volodina-napoles:2016:BEA11}.

\section{Related Work}

While this is the first year for a shared task focusing on sentence-level binary error identification, previous work and shared tasks have focused on the related tasks of intra-sentence identification and correction of errors. Until recently, standard hand-annotated grammatical error datasets were not available, complicating comparisons and limiting the choice of methods used. Given the lack of a large hand-annotated corpus at the time, \newcite{ParkEtAl-2011-UnsupervisedGEC} demonstrated the use of the EM algorithm for parameter learning of a noise model using error data without corrections, performing evaluation on a much smaller set of sentences hand-corrected by Amazon Mechanical Turk workers. 

More recent work has emerged as a result of a series of shared tasks, starting with the Helping Our Own (HOO) Pilot Shared Task run in 2011, which focused on a diverse set of errors in a small dataset \cite{DaleAndKilgarriff-2011-HOOPilot}, and the subsequent HOO 2012 Shared Task, which focused on the automated detection and correction of preposition and determiner errors \cite{DaleEtAl-2012-HOO2012PrepAndDetErrors}. The CoNLL-2013 Shared Task \cite{NgEtAl-2013-CoNLL-SharedTask2013}\footnote{\url{http://www.comp.nus.edu.sg/~nlp/conll13st.html}} focused on the correction of a limited set of five error types in essays by second-language learners of English at the National University of Singapore. The follow-up CoNLL-2014 Shared Task \cite{NgEtAl-2014-SharedTask2014}\footnote{\url{http://www.comp.nus.edu.sg/~nlp/conll14st.html}} focused on the full generation task of correcting all errors in essays by second-language learners.

As with machine translation (MT), evaluation of the full generation task is still an open research area, but a subsequent human evaluation ranked the output from the CoNLL-2014 Shared Task systems \cite{NapolesEtAl-2015-GLEU}. The system of \newcite{FeliceEtAl-2014-Hybrid} ranked highest, utilizing a combination of a rule-based system and phrase-based MT, with re-ranking via a large web-scale language model. Of the non-MT based approaches, the Illinois-Columbia system was a strong performer, combining several classifiers trained for specific types of errors \cite{RozovskayaEtAl-2014-IllinoisColumbiaClassifierSystem}.

\section{Models}

We use an end-to-end approach that does not have separate components for candidate generation or re-ranking that make use of hand-tuned rules or explicit syntax, nor do we employ separate classifiers for human-differentiated subsets of errors, unlike some previous work for the related task of grammatical error correction.

We next introduce two approaches for the task of sentence-level grammatical error identification: A binary classifier and a sequence-to-sequence model that is trained for correction but can also be used for identification as a side-effect.

\subsection{Baseline Convolutional Neural Net}

To establish a baseline, we follow past work that has shown strong performance with convolutional neural nets (CNNs) across various domains for sentence-level classification \cite{Kim-2014-CNN,ZhangAndWallace-2015-CNNGuide}. We utilize the one-layer CNN architecture of \newcite{Kim-2014-CNN} with the publicly available\footnote{\url{https://code.google.com/archive/p/word2vec/}} word vectors trained on the Google News dataset, which contains about 100 billion words \cite{MikolovEtAl-2013-Word2vec}. We experiment with keeping the word vectors static (\textsc{CNN-static}) and fine-tuning the vectors (\textsc{CNN-nonstatic}). The CNN models only have access to sentence-level labels and are not given correction-level annotations.

\subsection{Encoder-Decoder}\label{sec:Encoder-Decoder}
\begin{figure*}[t]
\center
\includegraphics[width=14cm]{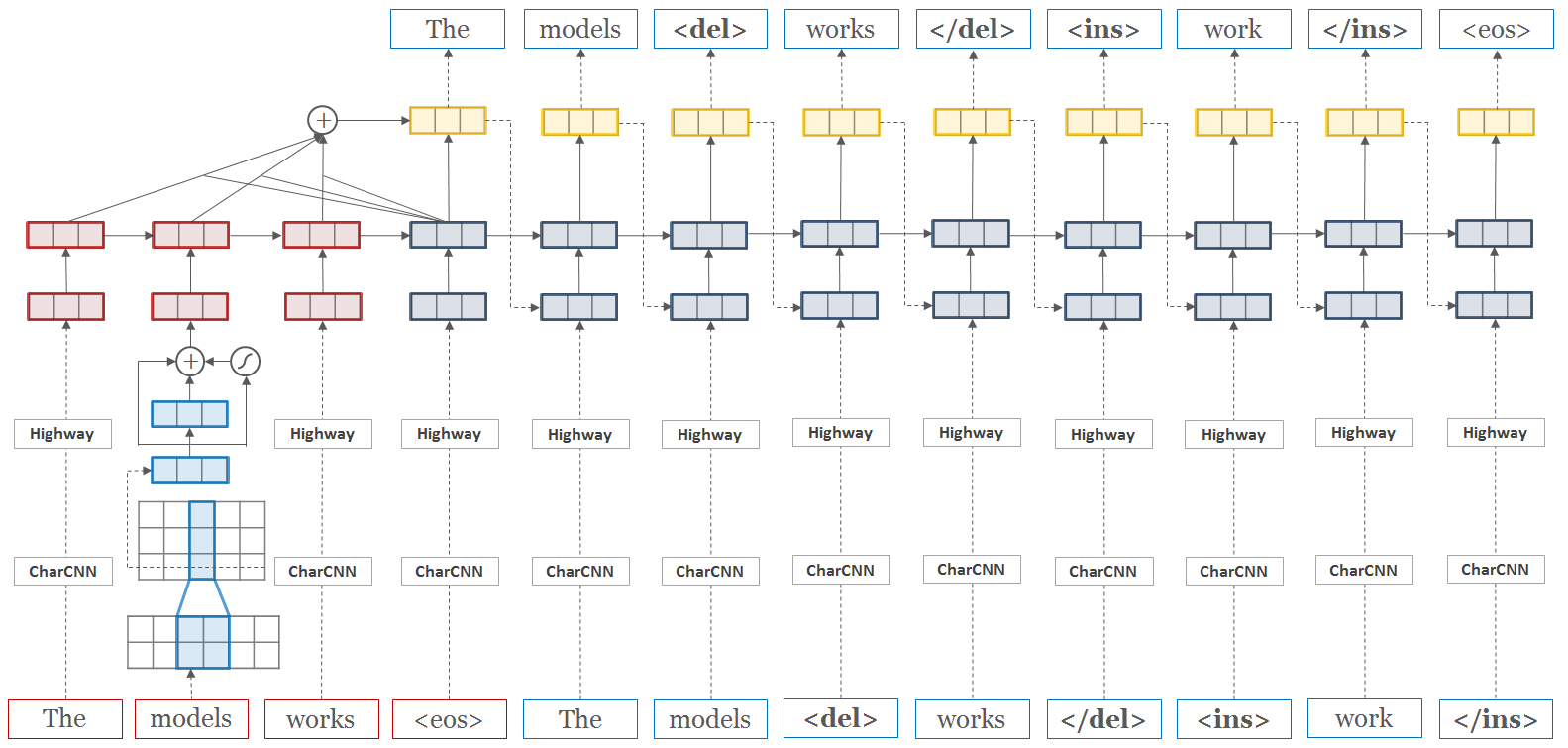}
\caption{\footnotesize An illustration of the \textsc{Char} model architecture translating an example source sentence into the corrected target with a single word replacement. A CNN (here, with three filters of width two) is applied over character embeddings to obtain a fixed dimensional representation of a word, which is given to a highway network (in light blue, above). Output from the highway network is used as input to a LSTM encoder/decoder. At each step of the decoder, its hidden state is interacted with the hidden states of the encoder to produce attention weights (for each word in the encoder), which are
used to obtain the context vector via a convex combination. The context vector is combined 
with the decoder hidden state through a one layer MLP (yellow), after which an affine 
transformation followed by a softmax is applied to obtain a distribution over the next word/tag.
The MLP layer (yellow) is used as additional input (via concatenation) for the next time step.
Generation continues until the $<$\texttt{eos}$>$ symbol is generated.}
\label{fig:architecture}
\end{figure*}

While it may seem more natural to utilize models trained for binary prediction, such as the aforementioned CNN, or for example, the recurrent network approach of \newcite{DaiAndLe-2015-Seq2SeqLSTMClassfierPreTraining}, we hypothesize that training at the lowest granularity of annotations may be useful for the task. We also suspect that the  
generation of corrections is of sufficient utility for end-users to further justify exploring models that produce corrections in addition to identification. We thus use the Shared Task as a means of assessing the utility of a full generation model for the binary prediction task.

We propose two encoder-decoder architectures for this task. Our word-based architecture (\textsc{Word}) is similar to that of \newcite{luong-pham-manning:2015:EMNLP}. Our character-based models (\textsc{Char}) still make predictions at the word-level, but use a CNN and a highway network over characters instead of word embeddings as the input to the encoder and decoder, as depicted in Figure~\ref{fig:architecture}. We follow past work \cite{SutskeverEtAl-2014-SequenceToSequence,luong-pham-manning:2015:EMNLP} in stacking multiple recurrent neural networks (RNNs), specifically Long Short-Term Memory (LSTM) \cite{Hochreiter1997_LSTM} networks, in both the encoder and decoder. 

Here, we model the probability of the target given the source, $p(\tvec\given\svec)$, with an \emph{encoder} neural network that summarizes the
source sequence and a \emph{decoder} neural network that generates a distribution over the target words and tags at each step given the source.

We start by describing the basic encoder and decoder architectures in terms of the \textsc{Word} model, and then we describe the \textsc{Char} model departures from \textsc{Word}. 

\paragraph{Encoder}

The encoder reads the source sentence and outputs a sequence of vectors, associated with each word in the sentence, which will be selectively accessed during decoding via a soft attentional mechanism. We use a LSTM
network to obtain the hidden states $\hvec^{s}_i \in \reals^{n}$ for each time step $i$,
\begin{equation*}
\hvec^{s}_i = \lstm(\hvec^{s}_{i-1}, \xvec_i^s).
\end{equation*}
For the \textsc{Word} models, $\xvec_i^s \in \reals^{m}$ is the word embedding for $s_i$, the $i$-th word in the source sentence. (The analogue for the \textsc{Char} models is discussed below.)
The output of the encoder is the sequence of hidden state vectors $[\hvec^{s}_1, \dots, \hvec^{s}_I]$.
The initial hidden state of the encoder is set to zero (i.e. $\hvec_0^s \leftarrow \mathbf{0}$).

\paragraph{Decoder}

The decoder is another LSTM that produces a distribution over the next target word/tag given 
the source vectors $[\hvec^{s}_1, \dots, \hvec^{s}_I]$ and the previously generated target words/tags 
$\tvec_{<j} = [t_1, \dots t_{j}]$. Let
\begin{equation*}
\hvec^{t}_j = \lstm(\hvec^{t}_{j-1}, \xvec_j^t)
\end{equation*}
be the summary of the target sentence up to the $j$-th word, where $\xvec_j^t$ is the word embedding
for $t_j$ in the \textsc{Word} models. The current target hidden state $\hvec_j^t$ is combined with each of the memory vectors
in the source to produce attention weights as follows,
\begin{align*}
u_{j,i} &= \hvec^t_j \cdot \Wvec_\alpha \hvec^s_i \\
\alpha_{j,i} &= \frac{\exp u_{j,i}}{\sum_{k\in[1,I]} \exp u_{j,k}} 
\end{align*}

The source vectors are multiplied with the respective attention weights, summed, and interacted
with the current decoder hidden state $\hvec_j^t$ to produce a \emph{context} vector $\cvec_j$,
\begin{align*}
\mathbf{v}_j &= \sum_{i \in [1,I]} \alpha_{j,i} \hvec_i^s \\
\cvec_j &= \tanh(\Wvec[\mathbf{v}_j;\hvec^t_j])
\end{align*}

The probability distribution over the next word/tag is given by applying an affine transformation
to $\cvec_j$ followed by a softmax,
\begin{equation*}
p(t_{j+1} \given \svec, \tvec_{<j}) = \softmax(\Uvec \cvec_j + \bvec)
\end{equation*}
Finally, as in \newcite{luong-pham-manning:2015:EMNLP}, we feed $\cvec_j$ as additional input to 
the decoder for the next time step by concatenating it with $\xvec_j^t$, so the decoder equation
 is modified to,
\begin{equation*}
\hvec^{t}_j = \lstm(\hvec^{t}_{j-1}, [\xvec_j^t; \cvec_{j-1}])
\end{equation*}
This allows the decoder to have knowledge of previous (soft) alignments at each time step.
The decoder hidden state is initialized with the final hidden state of the 
encoder (i.e. $\hvec_0^t \leftarrow \hvec_I^s$).

\paragraph{Character Convolutional Neural Network}\label{sec:CharCNN}
For the \textsc{Char} models, instead of a word embedding, our input for each word in the source/target 
sentence is an output from a character-level convolutional neural network (CharCNN) 
(depicted in Figure~\ref{fig:architecture}). Our character model closely follows that of \newcite{KimEtAl-2016-CharLM}.

Suppose word $s_i$ is composed of characters $[p_1, \dots, p_l]$. We concatenate the character 
embeddings to form the matrix $\mathbf{P}_i \in \reals^{c \times l}$, where the 
$k$-th column corresponds to the character embedding for $p_k$ (of dimension $c$).

We then apply a convolution between $\mathbf{P}_i$ and a {\em filter}  
$\mathbf{H} \in \reals^{c \times w}$ of width $w$, after which we
add a bias and apply a nonlinearity to obtain a {\em feature map} 
$\mathbf{f}_i \in \reals^{l-w+1}$. The $k$-th element of $\mathbf{f}_i$
is given by,
\begin{equation*}
\mathbf{f}_i[k] = \tanh(\langle \mathbf{P}_i[\ast, k:k+w-1], \mathbf{H} \rangle + b)
\end{equation*}
where $\langle \mathbf{A},\mathbf{B} \rangle = \mbox{Tr}(\mathbf{A}\mathbf{B}^T)
$ is the Frobenius inner product and $\mathbf{P}_i[\ast, k:k+w-1]$ is the
$k$-to-$(k+w-1)$-th column of $\mathbf{P}_i$. Finally, we take the {\em max-over-time}
\begin{equation*}
z_i = \max_k \mathbf{f}_i[k]
\end{equation*}
as the feature corresponding to filter $\mathbf{H}$. We use multiple filters 
$\mathbf{H}_1, \dots \mathbf{H}_h$ to
obtain a vector $\zvec_i \in \reals^h$ as the representation for a given source/target word or tag.
We have separate CharCNNs for the encoder and decoder.

\paragraph{Highway Network}\label{sec:Highway}
Instead of replacing the word embedding $\xvec_i$ with $\zvec_i$, we feed $\zvec_i$ through 
a {\em highway network} \cite{SrivastavaEtAl-2015-HighwayNets}. 
Whereas a multilayer perceptron produces a new set of features via the following 
transformation (given input $\zvec$),
\begin{equation*}
\hat{\zvec} = f(\Wvec\zvec + \bvec)
\end{equation*}
a highway network instead computes,
\begin{equation*}
\hat{\zvec} =  \mathbf{r} \odot f(\Wvec\zvec +  \bvec) +
               (\mathbf{1} - \mathbf{r}) \odot \zvec
\end{equation*}
where $f$ is a non-linearity (in our models, $\relu$), $\odot$ is the element-wise 
multiplication operator, and $\mathbf{r} = \sigma(\mathbf{W}_r\zvec + \bvec_r)$ 
and $\mathbf{1} - \mathbf{r}$ are called the {\em transform} and {\em carry} gates. 
We feed $\zvec_i$ into the highway network to obtain $\hat{\zvec}_i$, which is 
used to replace the input word embeddings in both the encoder and the decoder.

\paragraph{Inference}

Exact inference is computationally infeasible for the encoder-decoder models given the combinatorial explosion of possible output sequences, but we follow past work in NMT using beam search. We do not constrain the generation process of words outside insertion tags to words in the source, and each low-frequency holder token generated in the target sentence is replaced with the source token associated with the maximum attention weight. We use a beam size of 10 for all models, with the exception of one of the models in the final system combination, for which we use a beam of size 5, as noted in Section \ref{sec:Results}.

Note that this model generates corrections, but we are only interested in determining the existence of any error at the sentence-level. As such, after beam decoding, we simply check for whether there were any corrections in the target. However, we found that decoding in this way under-predicts sentence-level errors. It is therefore important to calibrate the weights associated with corrections, which we discuss in Section \ref{sec:ExperimentsTuning}.

\section{Experiments}

\subsection{Data}

The AESW task data differs from previous grammatical error datasets in terms of scale and genre. To the best of our knowledge, the AESW dataset is the first large-scale, publicly available professionally edited dataset of academic, scientific writing. The training set consists of 466,672 sentences with edits and 722,742 sentences without edits, and the development set contains 57,340 sentences with edits and 90,106 sentences without. The raw training and development datasets are provided as annotated sentences, $\mathbf{t}$, from which the $\mathbf{s}$ sequences may be deterministically derived. There are 143,802 sentences in the Shared Task test set with hidden gold labels, which serve directly as $\mathbf{s}$ sequences.

As part of pre-processing, we treat each sentence independently, discarding paragraph context (which sentences, if any, were present in the same paragraph) and domain information, which is a coarse grouping by the field of the original journal (Engineering, Computer Science, Mathematics, Chemistry, Physics, etc.). We generate Penn Treebank style tokenizations of the input. Case is maintained and digits are not replaced with holder symbols. The vocabulary is restricted to the 50,000 most common tokens, with remaining low frequency tokens replaced with a special $\mathsf{ \textless unk\textgreater}$ token. The
\textsc{Char} model can encode but not decode over open vocabularies and hence we do not have any $\mathsf{ \textless unk\textgreater}$ tokens on the source side of those models. For all of the encoder-decoder models, we replace the low-frequency target symbols during inference as discussed above in Section \ref{sec:Encoder-Decoder}. 

For development against the provided data with labels, we set aside a 10,000 sentence sample from the original development set for tuning, and use the remaining 137,446 sentences for validation\footnote{Note that the number of sentences in the final development set without labels posted on CodaLab (\url{http://codalab.org}) differed from that originally posted on the AESW 2016 Shared Task website with labels.}. The encoder-decoder models are given all 466,672 pairs of~$\mathbf{s}$~and~$\mathbf{t}$~sequences with edits, augmented with varying numbers of pairs without edits. The \textsc{Char+sample} and \textsc{Word+sample} models are given a random sample of 200,000 pairs without edits for a total of 666,672 pairs of~$\mathbf{s}$~and~$\mathbf{t}$~sequences. The \textsc{Char+all} and \textsc{Word+all} models are given all 722,742 sentences without edits for a total of 1,189,414 pairs of~$\mathbf{s}$~and~$\mathbf{t}$~sequences. For some of the final testing models, we also train with the development sentences. In these latter cases, all sequence pairs are used. In training all of the encoder-decoder models, as indicated in Section \ref{sec:Training}, we drop sentences exceeding 50 tokens in length.

We also experimented with creating corrected versions of sentences for the CNN. The binary CNN classifiers are given 1,656,086 single-sentence training examples, of which 722,742 are error-free examples (in which $\mathbf{s}=\mathbf{t}$), and the remaining examples are constructed by removing the tags from the annotated sentences, $\mathbf{t}$, to create tag-free examples that contain errors (466,672 instances) and additional error-free examples (466,672 instances). 
  
\subsection{Training} \label{sec:Training}

Training (along with testing) of all models was conducted on GPUs. Our models were implemented with the Torch\footnote{\url{http://torch.ch}} framework.

\paragraph{CNN} Architecture and training approaches were informed by past work in sentence-level classification using CNNs \cite{Kim-2014-CNN,ZhangAndWallace-2015-CNNGuide}. A limited grid search on the development set determined our use of filter windows of width 3, 4, and 5 and 1000 feature maps. We trained for 10 epochs. Training otherwise followed the approach of the correspondingly named \textsc{CNN-static} and \textsc{CNN-nonstatic} models of \newcite{Kim-2014-CNN}.

\paragraph{encoder-decoder} Initial parameter settings (including architecture decisions such as the number of layers and embedding and hidden state sizes) were informed by concurrent work in neural machine translation and existing work such as that of \newcite{SutskeverEtAl-2014-SequenceToSequence} and \newcite{luong-pham-manning:2015:EMNLP}. We used $4$-layer LSTMs with $1000$ hidden units in each layer. We trained for 14 epochs with a batch size of 64 and a maximum sequence length of 50. The parameters for the \textsc{Word} model were uniformly initialized in $[-0.1, 0.1]$, and those of the \textsc{Char} model were uniformly initialized in $[-0.05, 0.05]$. The $L_2$-normalized gradients were constrained to be $\le 5$. Our learning rate schedule started the learning rate at 1 and halved the learning rate after each epoch beyond epoch 10, or once the validation set perplexity no longer improved. The \textsc{Word} model used $1000$-dimensional word embeddings. For \textsc{Char}, the character embeddings were $25$-dimensional, the filter width was $6$, the number of feature maps was $1000$, and $2$ highway layers were used. The maximum word length was 35 characters for training \textsc{Char}. Note that we do not reverse the source ($\mathbf{s}$) sequences, unlike some previous NMT work. Following the work of \newcite{Zaremba14_rnn_regularization}, we employed dropout with a probability of $0.3$ between the LSTM layers.

Training both \textsc{Word} and \textsc{Char} on the training set took on the order of a few days using GPUs, with the former being more efficient than the latter. In practice, we used two GPUs for training \textsc{Char} due to memory requirements.

\subsection{Tuning}\label{sec:ExperimentsTuning}

Post-hoc tuning was performed on the 10k held-out portion of the development set. In terms of maximizing the $F_1$-score, this post-hoc tuning was important for these models, without which precision was high and recall was low. We leave to future work alternative approaches to this type of post-hoc tuning.

For the CNN models, after training, we tuned the decision boundary to maximize the $F_1$-score on the held-out tuning set. Analogously, for the encoder-decoder models, after training the models, we tuned the bias weights (given as input to the final softmax layer generating the words/tags distribution) associated with the four annotation tags via a simple grid search by iteratively running beam search on the tuning set. Due to the relatively high expense of decoding, we employed a coarse grid search in which the bias weights of the four annotation tags were uniformly varied.

\section{Results}\label{sec:Results}

\begin{figure}[t!]
    \centering
    \includegraphics{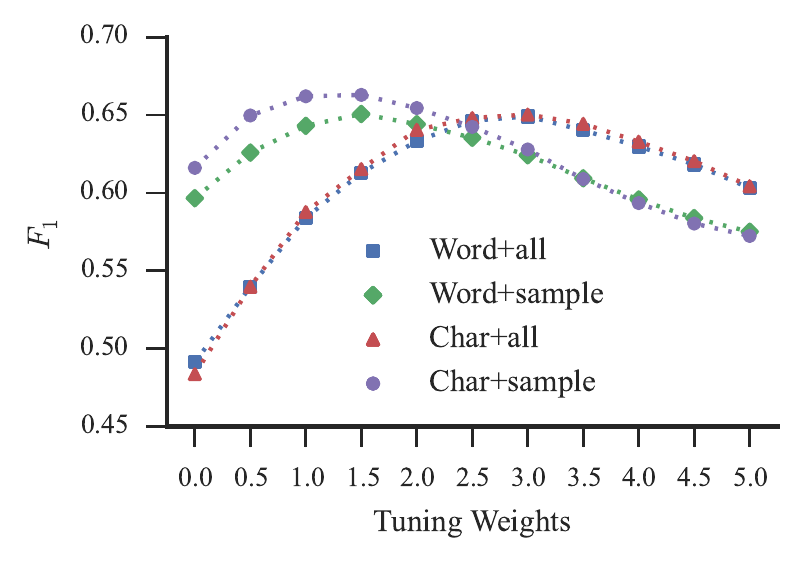}
    \caption{$F_1$ scores for varying values applied additively to the bias weights of the four annotation tags on the held-out 10k tuning subset.}
    \label{figure:f1graph}
\end{figure} 

\begin{figure}[t!]
    \centering
    \includegraphics{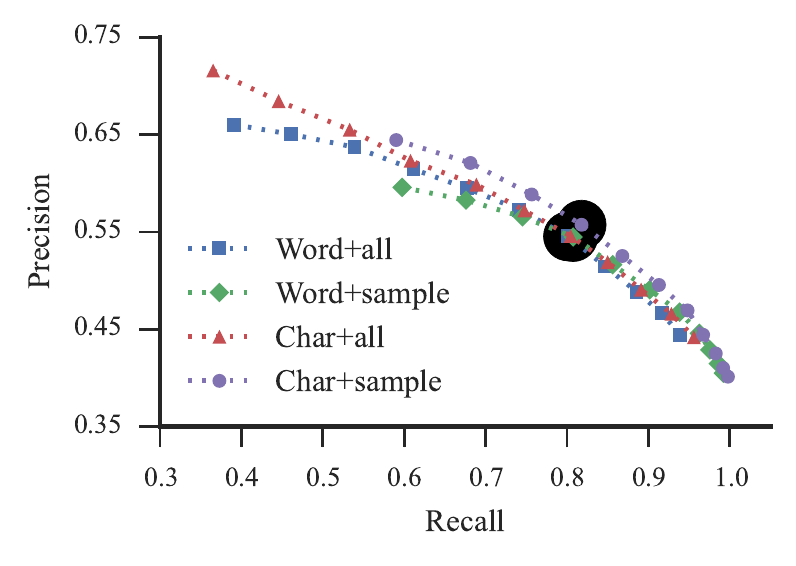}
    \caption{Precision vs. recall trade-off as the bias weights associated with the four annotation tags are varied on the held-out 10k tuning subset. The points yielding maximum $F_1$ scores are highlighted with black circles.}
    \label{figure:prgraph}
\end{figure}

\begin{table*}[ht!]
\centering
\footnotesize
\begin{tabular}{lcccc}
\toprule
Model & Data & Precision & Recall & $F_1$ \\
\midrule
\textsc{Random} & N/A & $0.3885$ & $0.4992$ & $0.4369$ \\
\midrule
\textsc{CNN-static} & Training+word2vec & $0.5349$ & $0.7586$ & $0.6274$ \\
\textsc{CNN-nonstatic} & Training+word2vec & $0.5365$ & $0.7758$ & $0.6343$ \\
\midrule
\textsc{Word+all} & Training & $0.5399$ & $0.7882$ & $0.6408$ \\
\textsc{Word+sample} & Training & $0.5394$ & $0.8024$ & $0.6451$ \\
\textsc{Char+all} & Training & $0.5400$ & $0.8048$ & $0.6463$ \\
\textsc{Char+sample} & Training & $0.5526$ & $0.8126$ & $0.6579$ \\
\bottomrule
\end{tabular}
\caption{Experimental results on the development set excluding the held-out 10k tuning subset.}
\label{tab:dev-results}
\end{table*}

\begin{table*}[ht!]
\centering
\footnotesize
\begin{tabular}{lcccc}
\toprule
Model & Data & Precision & Recall & $F_1$ \\
\midrule
\textsc{Random} & N/A & $0.3921$ & $0.5981$ & $0.4736$ \\
\textsc{Word+all} & Training & $0.5343$ & $0.7577$ & $0.6267$ \\
\textsc{Word+sample} & Training & $0.5335$ & $0.7699$ & $0.6303$ \\
\textsc{Char+all} & Training & $0.5351$ & $0.7749$ & $0.6330$ \\
\textsc{Char+sample} & Training & $0.5469$ & $0.7803$ & $0.6431$ \\
\bottomrule
\end{tabular}
\caption{Results on the official development set. Here, \textsc{Random} was provided by the Shared Task organizers.}
\label{tab:codalabdev-results}
\end{table*}

\begin{table*}[ht!]
\centering
\footnotesize
\begin{tabular}{lcccc}
\toprule
Model & Data & Precision & Recall & $F_1$ \\
\midrule
\textsc{Random} & N/A & $0.3607$ & $0.6004$ & $0.4507$ \\
\midrule
\textsc{Knowlet} & -- & $0.6241$ & $0.3685$ & $0.4634$ \\
\textsc{NTNU-YZU} & -- & $0.6717$ & $0.3805$ & $0.4858$ \\
\textsc{HITS} & -- & $0.3765$ & $0.948$ & $0.5389$ \\
\textsc{UW-SU} & -- & $0.4145$ & $0.8201$ & $0.5507$ \\
\textsc{NTNU-YZU} & -- & $0.5025$ & $0.7785$ & $0.6108$ \\
\midrule
\textsc{Char+sample} & Training & $0.5112$ & $0.7841$ & $0.6189$ \\
\textsc{Combination} & Training+Dev+word2vec & $0.5444$ & $0.7413$ & $0.6278$ \\
\bottomrule
\end{tabular}
\caption{\footnotesize Final submitted results on the Shared Task test set. \textsc{Combination} was our final submitted system. \textsc{Random} was provided by the Shared Task organizers. For comparison, we have included the other team submissions from National Taiwan Normal University and Yuan Ze University (\textsc{NTNU-YZU}), the University of Washington and Stanford University (\textsc{UW-SU}), HITS (\textsc{HITS}), and Knowlet (\textsc{Knowlet}). Teams were allowed to designate up to two final submissions. (The \textsc{Char} model trained on the combined training and development set had not finished training by the Shared Task deadline. As such, it was not submitted, but the partially trained model was included in \textsc{Combination}.)}
\label{tab:test-results}
\end{table*}

Results on the development set, excluding the 10k tuning set, appear in Table \ref{tab:dev-results}. Here (and elsewhere) \textsc{Random} is the result of randomly assigning a sentence to one of the binary classes. For the CNN classifiers, fine-tuning the word2vec embeddings improves performance. The encoder-decoder models improve over the CNN classifiers, even though the latter are provided with additional data (via word2vec). The character-based models yield tangible improvements over the word-based models.

For consistency here, we kept the beam size at 10 across models, but subsequent analysis revealed that increasing the beam from 5 to 10 had a negligible effect on overall performance.

Tuning results appear in Figures \ref{figure:f1graph} and \ref{figure:prgraph}, illustrating the importance of adjusting the bias weights associated with the annotation tags in balancing precision and recall to maximize the $F_1$ score. The models trained on all sequence pairs without edits, \textsc{Char+all} and \textsc{Word+all}, perform particularly poorly without tuning these bias weights, yielding $F_1$ scores near that of \textsc{Random} before tuning, which corresponds to a weight of 0.0 in Figure \ref{figure:f1graph}. 

The official development set posted on CodaLab differed slightly from the original development set provided with labels, so we include those results in Table \ref{tab:codalabdev-results} for the encoder-decoder models. Here, evaluation is performed on the CodaLab server, as with the final test submission. The overall relative performance pattern is similar to that of the original development set. 

A comparison of our results with other shared task submissions appears in Table \ref{tab:test-results}. (Teams were allowed to submit up to two results.) Our submission, \textsc{Combination} was a simple majority vote at the system level (for each test sentence) of 5 models\footnote{The choice of models was limited to those that were trained and tuned in time for the Shared Task deadline.}: (1) a \textsc{CNN-nonstatic} model trained with the concatenation of the training and development sets (and using word2vec); (2) a \textsc{Word} model trained on all sequence pairs in the training and development sets with a beam size of 10 for decoding; (3,4) a \textsc{Char+sample} model trained on the training set, decoding the test set twice, each time with different weight biases (the two highest performing via the grid search over the tuning set) with a beam size of 10; and (5) a \textsc{Char} model trained on all sequence pairs in the training and development sets, with training suspended at epoch 9 (out of 14) and a beam size of 5 to meet the Shared Task deadline. For reference, we also include the CodaLab evaluation for just the \textsc{Char+sample} model trained on the training set with a beam size of 10, with the bias weights being those that generated the highest $F_1$-score on the 10k tuning set.  

\section{Discussion}
 
Of particular interest, the \textsc{Char+sample} model performs well, both in terms of performance on the test set relative to other submissions, as well as on the development set relative to the \textsc{Word} models and the CNN classifiers. It is possible this is due to the ability of the \textsc{Char} models to capture some types of orthographic errors. 

The empirical results suggest that simply adding additional already correct source-target pairs when training the encoder-decoder models may not boost performance, ceteris paribus, as seen in comparing the performance of \textsc{Char+sample} vs \textsc{Word+sample}, and \textsc{Char+all} vs \textsc{Word+all}. We leave to future work alternative approaches for introducing additional correct (target) sentences, as has been examined for neural machine translation models \cite{SennrichEtAl-2015-BacktranslationForNMT,GulcehreEtAl-2015-MonolingualDataForNMT}.

Our results provide initial evidence to support the hypothesis that training at the lowest granularity of annotation is a more efficient use of data than training against the binary label. In future work, we plan to compare against sentence classification using LSTMs \cite{DaiAndLe-2015-Seq2SeqLSTMClassfierPreTraining} and convolutional models that use correction-level annotations.

Another benefit of the encoder-decoder models is that they can be used to generate corrections (and identify locations of intra-sentence errors) for end-users. However, the added generation capabilities of the encoder-decoder models comes at the expense of considerably longer training and testing times compared to the CNN classifiers.

We found that post-hoc tuning provides a straightforward means of tuning the precision-recall trade-off for these models, and we speculate (but leave to future work for investigation) that in practice, end-users might prefer greater emphasis placed on precision over recall.

\section{Conclusion}

We have presented our submission to the AESW 2016 Shared Task, suggesting, in particular, the utility of a neural attention-based model for sentence-level grammatical error identification. Our models do not make use of hand-tuned rules, are not trained with explicit syntactic annotations, and do not make use of individuals classifiers designed for human-designated subsets of errors. 

For the encoder-decoder models, modeling at the sub-word level was beneficial, even though predictions were still made at the word level. It would be of interest to push this further to eliminate the need for an initial tokenization step, in order to generalize the approach to other languages, such as Chinese and Japanese. 

We plan to examine alternative approaches for training with additional correct (target) sentences. Inducing artificial errors to generate more incorrect (source) sentences is also a direction we intend to pursue.

We leave for future work an analysis of the generation quality of our encoder-decoder models on the AESW dataset and the CoNLL-2014 Shared Task data, as well as user studies to assess whether performance is sufficient in practice to be useful, including the utility of correction vs. identification.

We consider this to be just the beginning of the development of data-driven support tools for writers, and many areas remain to be explored.

\section*{Acknowledgments}

We would like to thank the organizers of the Shared Task for coordinating the task and making the unique AESW dataset available for research purposes. The Institute for Quantitative Social Science (IQSS) and the Harvard Initiative for Learning and Teaching (HILT) supported earlier, related research that led to our participation in the Shared Task. Jeffrey Ling graciously contributed a torch-based CNN implementation of \newcite{Kim-2014-CNN}. 

\balance
\bibliography{aesw2016}
\bibliographystyle{naaclhlt2016}


\end{document}